%% file: main.tex
\pdfoutput=1

\documentclass[11pt]{article}

\usepackage[final]{acl}

\usepackage{times}
\usepackage{latexsym}

\usepackage[T1]{fontenc}

\usepackage[utf8]{inputenc}

\usepackage{microtype}

\usepackage{inconsolata}

\usepackage{graphicx}

\usepackage{hyperref}  
\usepackage{amsmath}   
\usepackage{amssymb}   
\usepackage{booktabs}  
\usepackage{xcolor}    
\usepackage{caption}   
\usepackage{subcaption} 
\usepackage{algorithm} 
\usepackage{algpseudocode} 
\usepackage{listings}  
\usepackage{url}       
\usepackage{cleveref}  

\usepackage{multirow}
\usepackage{multicol}

\usepackage{adjustbox}

\PassOptionsToPackage{dvipsnames,table}{xcolor}

\input{macros}

\title{Scaling Up Membership Inference:\\When and How Attacks Succeed on Large Language Models}

\author{
 \textbf{Haritz Puerto\textsuperscript{1,2,\thanks{Work done during an internship at Parameter Lab.}}},
 \textbf{Martin Gubri\textsuperscript{1}},
 \textbf{Sangdoo Yun\textsuperscript{3}},
 \textbf{Seong Joon Oh\textsuperscript{1,4,5}},
\\
\\
 \textsuperscript{1}Parameter Lab,
 \textsuperscript{2}Ubiquitous Knowledge Processing Lab, Technical University of Darmstadt,\\
 \textsuperscript{3}NAVER AI Lab,
 \textsuperscript{4}University of Tübingen,
 \textsuperscript{5}Tübingen AI Center
\\
 \small{
   \href{haritz.puerto@tu-darmstadt.de@}{haritz.puerto@tu-darmstadt.de}
 }
}

\begin{document}
\maketitle
\begin{abstract}
Membership inference attacks (MIA) attempt to verify the membership of a given data sample in the training set for a model. 
MIA has become relevant in recent years, following the rapid development of large language models (LLM).
Many are concerned about the usage of copyrighted materials for training them and call for methods for detecting such usage. 
However, recent research has largely concluded that current MIA methods do not work on LLMs.
Even when they seem to work, it is usually because of the ill-designed experimental setup where other shortcut features enable ``cheating.'' 
In this work, we argue that MIA still works on LLMs, but only when multiple documents are presented for testing.
We construct new benchmarks that measure the MIA performances at a continuous scale of data samples, from sentences (n-grams) to a collection of documents (multiple chunks of tokens). 
To validate the efficacy of current MIA approaches at greater scales, we adapt a recent work on Dataset Inference (DI) for the task of binary membership detection that aggregates paragraph-level MIA features to enable MIA at document and collection of documents level.
This baseline achieves the first successful MIA on pre-trained and fine-tuned LLMs.\footnote{Our code is available at \href{https://github.com/parameterlab/mia-scaling}{https://github.com/parameterlab/mia-scaling}}

\end{abstract}

\section{Introduction}
\input{sections/1-intro}

\section{Background \& Related Work}

\input{sections/2-related}

\section{MIA Evaluation}
\label{sec:evaluation}
\input{sections/3-evaluation}

\section{Method}
\label{sec:method}
\input{sections/4-methods}

\section{Experiments}
\label{sec:experiments}
\input{sections/5-experiments}

\section{Conclusion}
\label{sec:conclusion}
\input{sections/6-conclusion}

\section*{Limitations}
\input{sections/limitations}

\section*{Ethics and Broader Impact Statement}
\input{sections/ethics}

\section*{Acknowledgements}

This work was supported by the NAVER corporation.

\bibliography{custom}

\appendix

\section{Experimental Settings}
\label{app:setup}

\paragraph{The Pile Data Splits.} We conduct our experiments on 13 out of the 22 subsets available in The Pile dataset. We exclude the nine subsets that contain less than 3000 documents in the validation and test splits (BookCorpus2, Books3, Ubuntu IRC, EuroParl, Enron Emails, OpenSubtitles, PhilPapers, Youtube Subtitles, and Gutenberg) because our evaluation needs enough non-members documents. We did not evaluate document membership on the PubMed Abstracts subset because the documents contain less than 512 tokens, and thus, we cannot aggregate paragraphs. To obtain the sentences for sentence-level MIA, we used the first 2048 tokens of each document and tokenized them using the NLTK sentence tokenizer \citep{nltk}. 

\paragraph{Implementation.} We use the MIA implementations from \citet{shi2024detecting}. We run all our experiments on five random seeds. We report the average AUROC with its standard deviation. To train the linear mapping that aggregates multiple MIA signals in the first stage of our method (\Cref{sec:method}), we use 1k members and 1k non-members from the known partition.

\paragraph{Sentence-Level MIA.} Since our base unit for aggregations is the sentence, sentence-level MIA does not use the second stage of our method (i.e., the aggregation part in \Cref{sec:method}), and instead only uses the first stage. Hence, it consists of learning a linear classifier that aggregates all the MIA methods and assigns a score to the input sentence.

\paragraph{Continual Learning Fine-tuning.} For the experiments on continual learning, we train Pythia 2.8B twice, once on the Wikipedia documents of the validation set and once on the Github documents of the validation set. We train the model with LoRa \citep{hu2021lora}, using the PEFT library \citep{peft}. Table \ref{table:ft-hps} reports the generic and LoRa-specific hyperparameters. We apply MIA using the validation set of The Pile as the members and the test set as the non-members. 

\paragraph{Hardware.} All experiments are run using PyTorch 1.13, Tesla V100-PCIE-32GB GPUs, CUDA 12.1 and Ubuntu 20.04.4 using NSML for MLAAS platform \citep{NSML}.

\input{tables/finetuning_hps}


\input{tables/u-test}

\section{MIA Benchmmark}
\label{appendix:mia_benchmark}
\Cref{table:mia_benchmark} shows the results of the membership inference attacks (MIA) across all data levels. We can observe how the \textit{collection} is the level with the highest proportion of successful MIA, while \textit{sentence} level does not show any. Collection MIA can only achieve results higher than 0.6 if paragraph MIA is significantly better than random. Document MIA further needs long enough documents.
\input{tables/mia_benchmark}

\section{T-Test vs. U-Test for Collection MIA}
In our experiments, we use t-tests for collection MIA and u-tests for document MIA because while the first one typically uses more than 30 samples (paragraphs) for document, and thus, fulfills the assumptions of t-tests, the second one does not. However, it would be possible to run collection MIA with u-test too. \Cref{tab:u-test} shows the comparison between t-test and u-test for collection MIA. We observe that u-tests yield better results than t-test. We hypothesize it might occurred because the MIA scores are not normally distributed, which is a general assumption of t-test, although not needed for samples larger than 30.

\section{Additional Figures}
\label{app:figures}
\begin{figure}[t]
    \centering
    \includegraphics[width=0.45\textwidth]{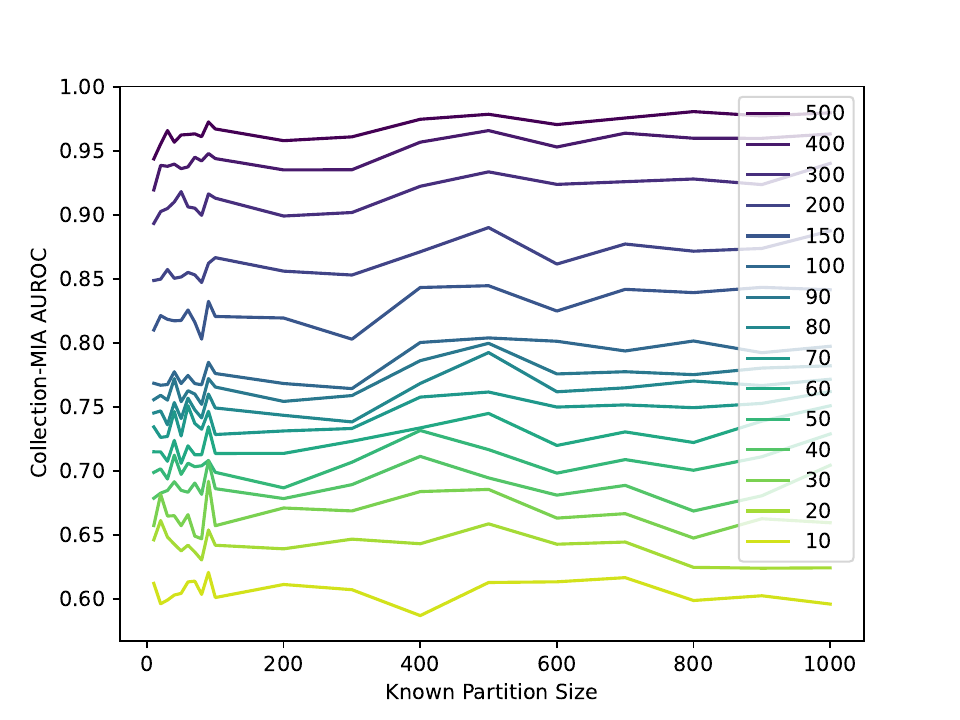}
    \caption{Increasing the size of the known partition does not increase Collection-MIA AUROC for any collection size.}
    \label{fig:known_size}
\end{figure}

\begin{figure}[t]
\centering
\includegraphics[width=0.45\textwidth]{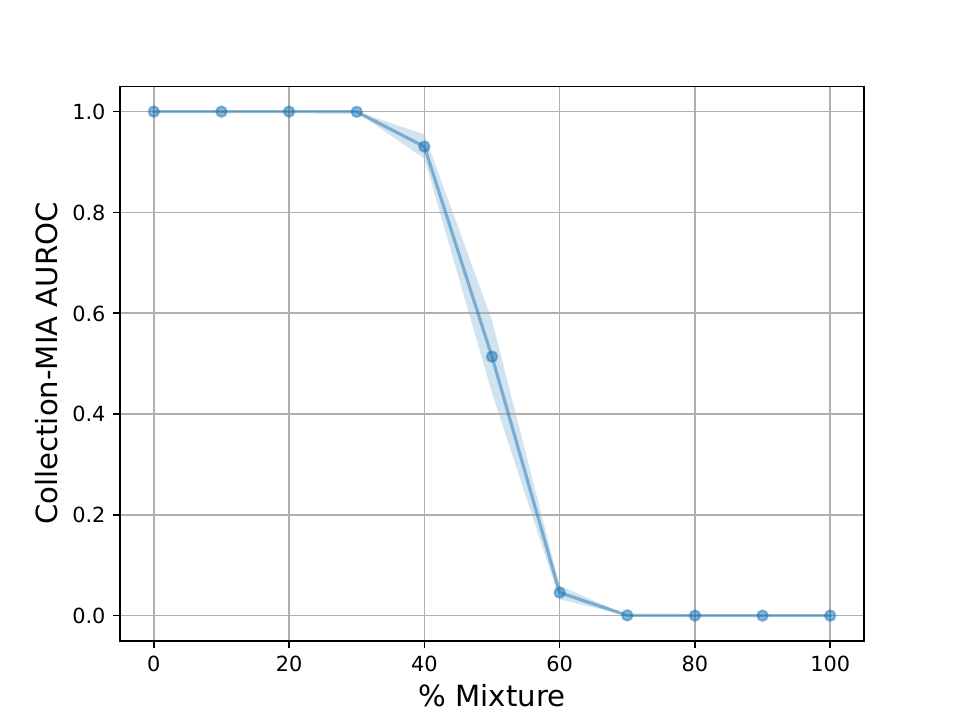}
\caption{Collection-MIA on CoT-based fine-tuned Phi 2 where the collections contain a \% of mix labels.}
\label{fig:dataset_mi_mix_ft}
\end{figure}

\begin{figure*}[t]
\centering
\includegraphics[width=\textwidth]{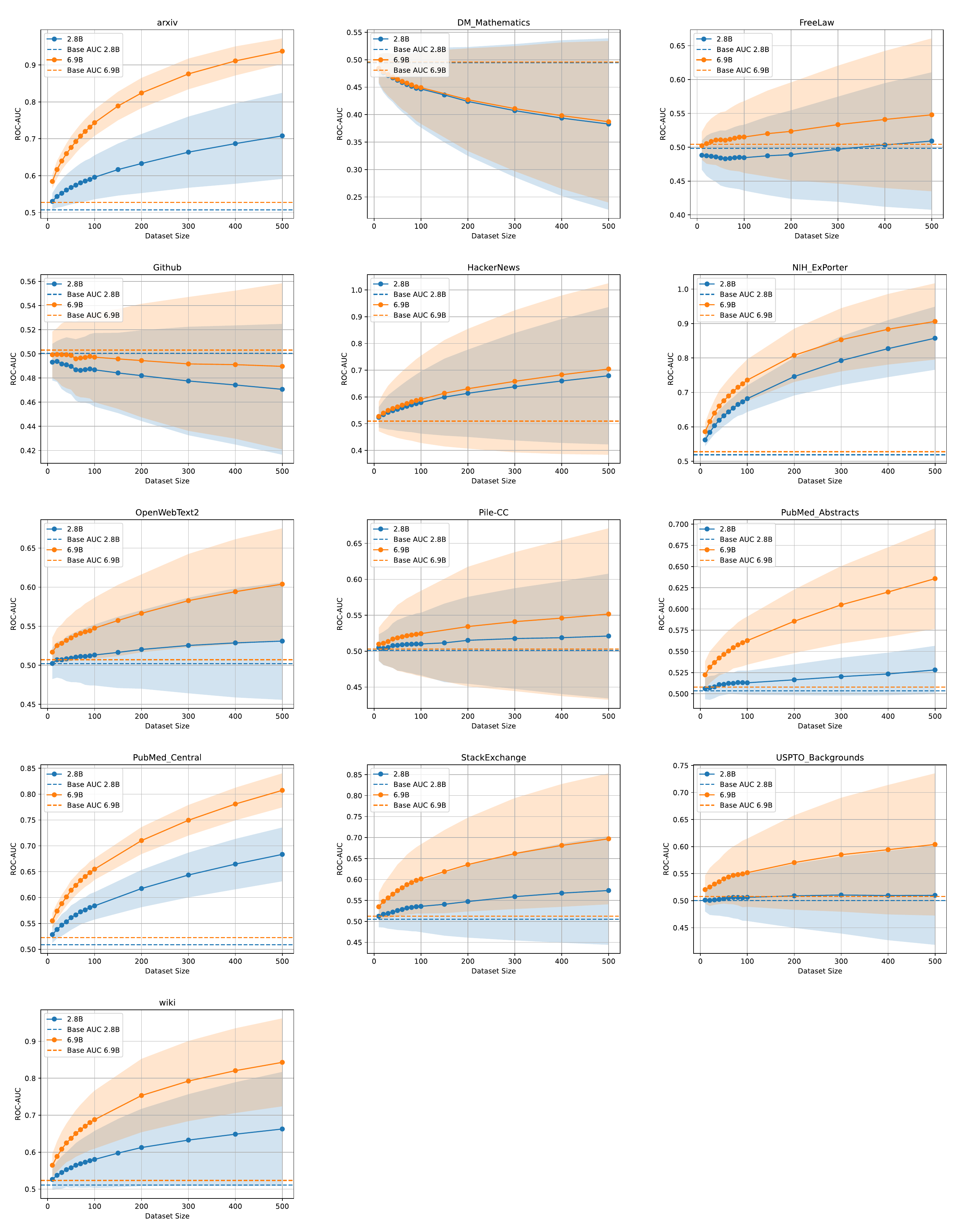}
\caption{Collection MIA on pretrained Pythia 2.8B and 6.9B}
\label{fig:appendix_dataset_mia}
\end{figure*}

\begin{figure*}[t]
\centering
\includegraphics[width=\textwidth]{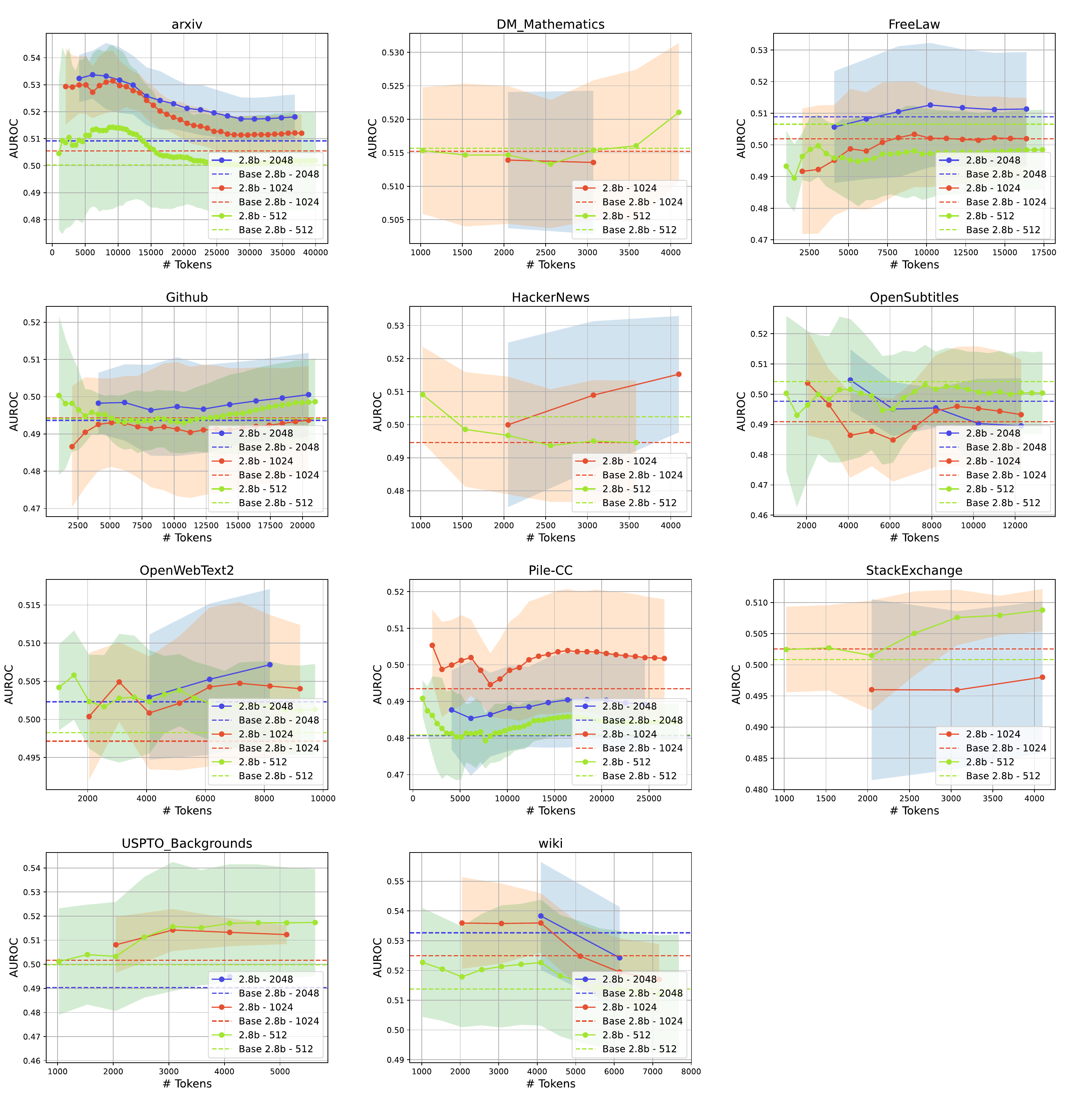}
\caption{Document MIA on pretrained Pythia 2.8B}
\label{fig:appendix_doc_mia_2B}
\end{figure*}

\begin{figure*}[t]
\centering
\includegraphics[width=\textwidth]{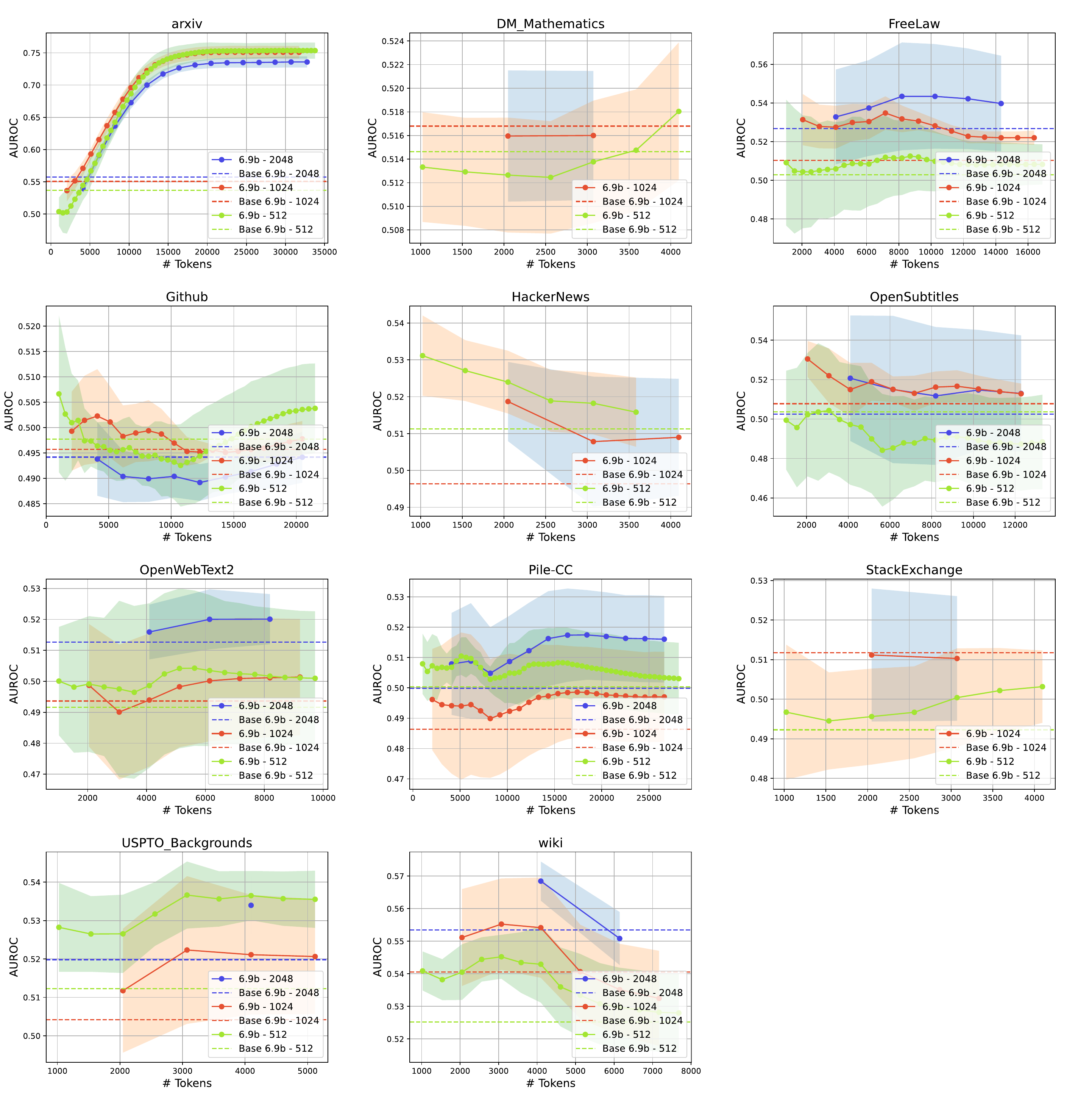}
\caption{Document MIA on pretrained Pythia 6.9B}
\label{fig:appendix_doc_mia_6B}
\end{figure*}

\begin{figure*}[t]
\centering
\includegraphics[width=\textwidth]{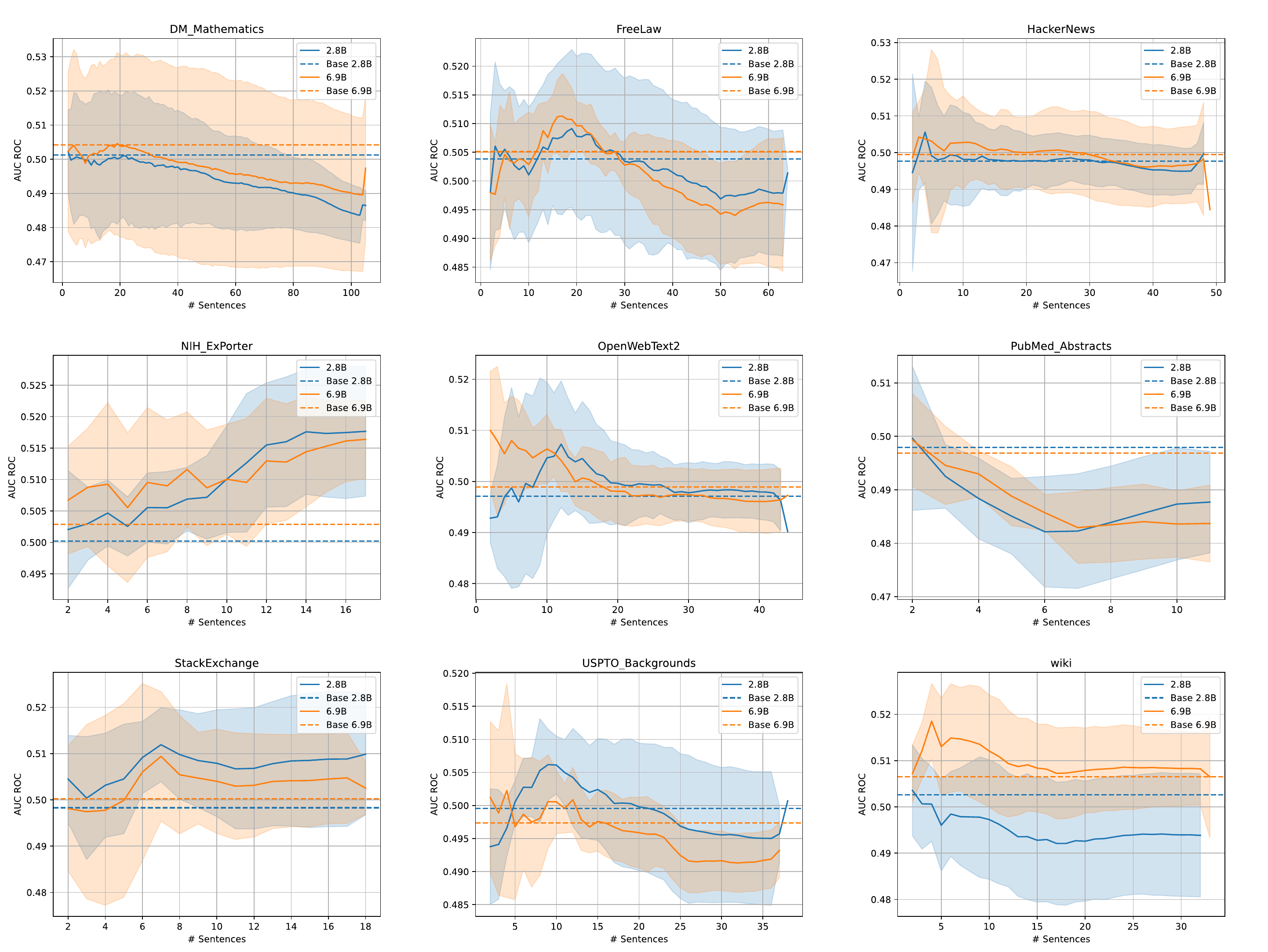}
\caption{Sent MIA on pretrained Pythia 2.8B and 6.9B}
\label{fig:appendix_pretrain_sent}
\end{figure*}

\begin{figure*}[t]
\centering
\includegraphics[width=0.8\textwidth]{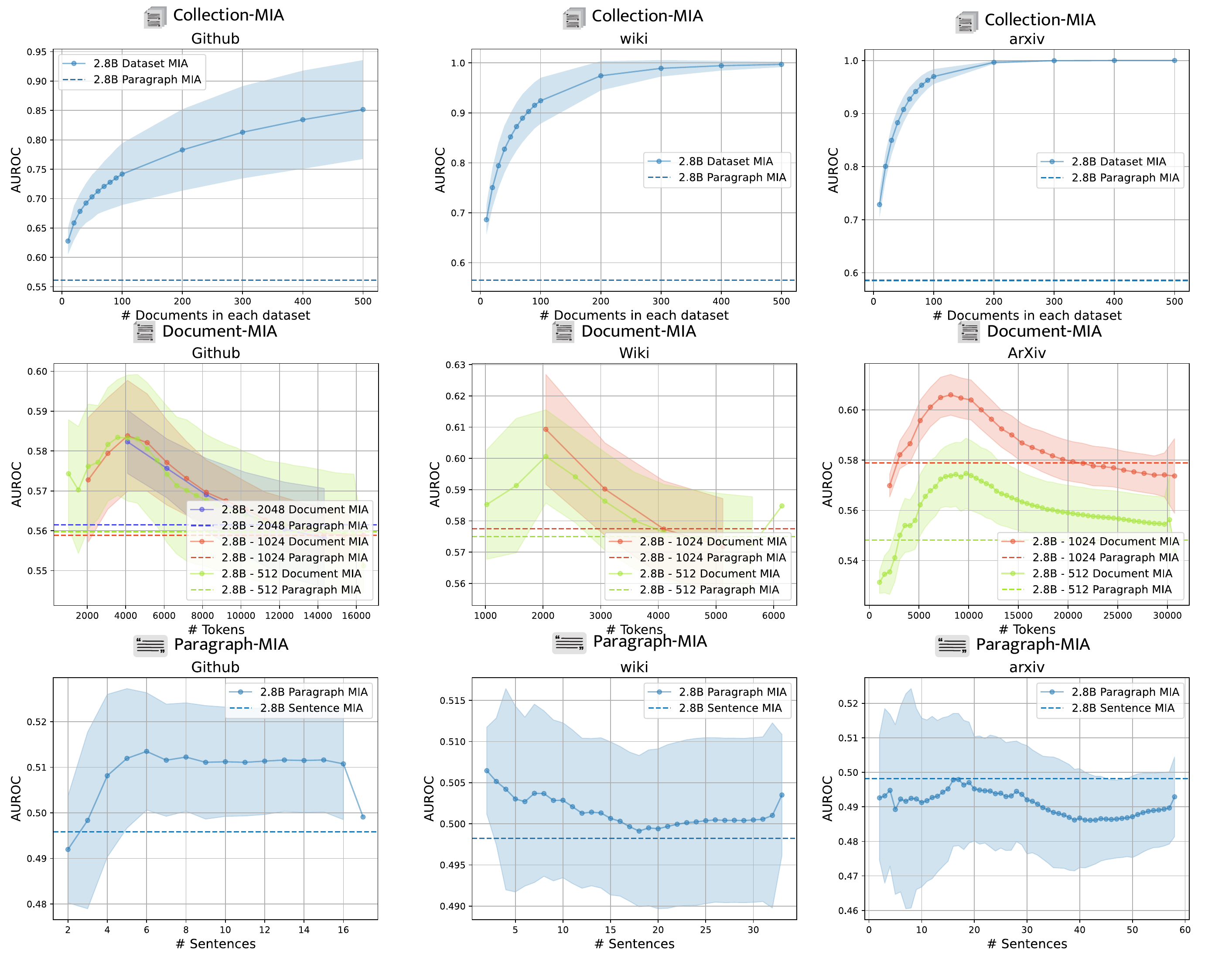}
\caption{MIA across all data scales Pythia 2.8B trained with continual learning.}
\label{fig:appendix_doc_mia_2B_cl}
\end{figure*}

In this section we include all figures that did not fit on the main paper. \Cref{fig:known_size} shows that there is no correlation between the known partition used for the statistical comparison and the AUROC. \Cref{fig:dataset_mi_mix_ft} shows the AUROC of collection-level MIA where the collections contain a \% of contaminated labels. Lastly, \Cref{fig:appendix_dataset_mia}, \ref{fig:appendix_doc_mia_2B}, \ref{fig:appendix_doc_mia_6B}, \ref{fig:appendix_pretrain_sent}, \ref{fig:appendix_doc_mia_2B_cl} show MIA performance across text granularities.

\section{Additional Tables}
\Cref{tab:token-metrics} presents statistics of the number of tokens across text granuliarities.

\input{figures/tab_nb_tokens_datasets}

\end{document}

%% file: macros.tex
\usepackage{xspace}
\usepackage[dvipsnames]{xcolor}

\newcommand{\ifprecedingtext}[1]{\ifvmode\relax\else#1\fi}

\definecolor{BrickRed}{RGB}{203,65,84}

\newenvironment{redenv}{
    \color{BrickRed}
}{
    \ignorespacesafterend
}

\newenvironment{blueenv}{
    \color{blue}
}{
    \ignorespacesafterend
}

\newenvironment{orangeenv}{
    \color{orange}
}{
    \ignorespacesafterend
}

\newenvironment{purpleenv}{
    \color{black}
}{
    \ignorespacesafterend
}

\newenvironment{oliveenv}{
    \color{olive}
}{
    \ignorespacesafterend
}



%% file: sections/1-intro.tex
Large language models (LLMs) are trained on vast datasets, which providers typically keep private. Data owners are concerned that their copyrighted data might be used in LLM training without explicit consent. Membership Inference Attacks (MIAs) attempt to determine if a specific data sample was used to train a model \citep{mia}. These methods are now being applied to LLMs to address the question of potential data misuse \citep{shi2024detecting, meeus2024did, zhang_min-k_2024,wang_con-recall_2024, xie2024recall}.

\begin{figure}[t]
    \centering
    \includegraphics[width=\linewidth]{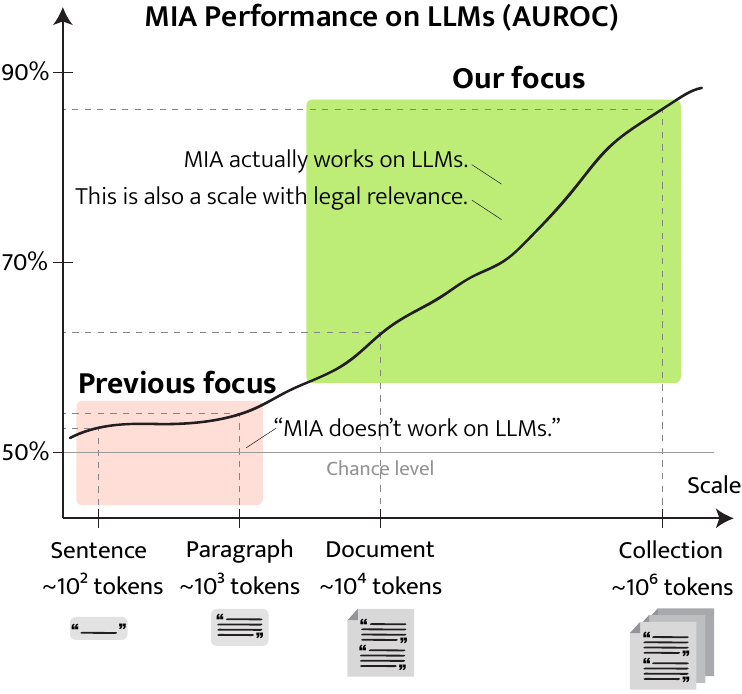}
    \vspace{-1em}
    \caption{\textbf{Focusing on the Right Scale.} MIA has traditionally been considered ineffective for LLMs. However, we argue that MIA remains effective for LLMs when applied at a much larger scale, considering significantly longer token sequences. This \textit{large-scale MIA} is also practically and legally relevant, as copyright is often determined at the document level.}
    \label{fig:intro}
\end{figure}
The application of MIAs on pretrained LLMs has faced discouraging results so far.  
Despite the initial optimism from the works of \citep{shi2024detecting, meeus2024did, carlini2021extracting, mattern_membership_2023}, \citet{duan2024membership,das_blind_2024,maini_llm_2024, meeus_inherent_2024} show later that current methods only achieve near random-guessing performance. 
Before their work, it was common practice to select true non-members from documents created after the LLM’s \textit{cut-off date}. They showed that this approach allowed MIA methods to exploit temporal cues, rather than identifying membership through the model's inherent response characteristics. \citet{duan2024membership,das_blind_2024,maini_llm_2024} introduced a new evaluation method based on an independent, identically distributed (IID) split between true members and non-members. Since adopting this method, no further MIA studies on LLMs have shown performance significantly better than random. Reported membership detection has remained below 60\% AUROC, close to the random-chance level of 50\% AUROC \citep{duan2024membership,das_blind_2024,maini_llm_2024, xie2024recall, zhang_min-k_2024}.

We argue that MIA can still be effective on LLMs, provided it is applied to much longer token sequences than previously considered. Earlier MIA approaches often focused on short sequences of tokens, typically ranging from 128 to 256 tokens \citep{xie2024recall, duan2024membership, shi2024detecting, Wang2024ConReCallDP}. This use of n-grams faced criticism because of the significant overlap between members and non-members, making the MIA problem poorly defined \citep{duan2024membership}.
Some later studies suggested analysing entire documents instead of n-grams \citep{shi2024detecting, meeus_inherent_2024}, but even then, performance remained near random \citep{meeus_inherent_2024}. 

In this work, we demonstrate that MIA approaches begin to show meaningful performance only when applied to much longer token sequences, such as 10K tokens.
To show this, we introduce four scales of token sequences: sentences, paragraphs, documents, and datasets, as shown in \Cref{fig:intro}.
We propose MIA evaluation protocols and benchmarks for the binary detection of training data samples given at four different scales.
As a baseline, we adapt the Dataset Inference \citep{maini_llm_2024} method for the MIA task at multiple scales.
As a result, our aggregation-based MIA demonstrates significant performance improvements for document sets, achieving AUROC scores of 80\% or higher.

To explore additional scenarios, we investigate the performance of MIA on fine-tuning data. Our findings show that continual learning MIA is effective collection of documents (AUROC > 88\%), while CoT-based fine-tuning works at both the sentence and collection levels. 

Our contributions are summarized as:
\begin{enumerate}
    \item We introduce a novel evaluation benchmark and protocol for MIA, covering multiple scales of token sequences. 
    \item We extend and adapt the first successful MIA \citet{maini_llm_2024} to any data scale, allowing us to conduct a comprehensive analysis of MIA performance in LLMs.
    \item We provide additional MIA benchmarks for various LLM fine-tuning scenarios, demonstrating that our method achieves even stronger performance in these contexts.
\end{enumerate}

%% file: sections/2-related.tex
Membership inference attacks (MIA) aims to prove that a certain data sample belongs to the training set of a model. 
\citet{yeom2018privacy} hypothesize that members have a lower loss (perplexity for LLMs) than non-members and based on this propose to use the loss to infer membership. \citet{carlini2021extracting} build on top of it and propose to use the ratio of the perplexity and the zlib entropy. \citet{shi2024detecting} propose to compute the average log-likelihood of the tokens with the lowest probabilities based on the assumption that members would have higher probabilities than non-members.

Since most LLMs do not have training-test set splits, initial benchmarks and recent works use the knowledge cut-off date to define members and non-members \citep{shi2024detecting, meeus2024did}. Works using these evaluation benchmarks show positive results \citep{shi2024detecting, meeus2024did, zhang_min-k_2024,wang_con-recall_2024, xie2024recall}. However, \citet{duan2024membership,das_blind_2024,maini_llm_2024} showed that the evaluation method based on cut-offs is flawed as these cut-offs introduce temporal biases that are easily captured by a bag of words. Therefore, they proposed to use LLMs trained on datasets that contain a random train-test split for members and non-members, like The Pile dataset \citep{pile}. In this setup, however, MIA only achieves performance barely above random guessing.

\citet{maini_llm_2024} show that aggregating MIA scores across multiple documents can yield successful dataset-level MIA. However, their work does not provide standard performance metrics like AUROC, leaving the precise effectiveness of their method unclear compared to standard paragraph-level MIA. In this study, we extend their evaluation by applying bootstrapping to compute AUROC for collection of documents and adapt their method to any data scale, allowing us to conduct a comprehensive analysis of MIA performance in LLMs. 

MIA performance across data scales in LLMs has not yet been explored despite its importance for copyright disputes. Current lawsuits against LLM developers primarily concern the use of entire documents in training sets \citep{harvard2024nyt}. However, most MIA methods focus on short paragraphs of up to 128 tokens \citep{xie2024recall}, 200 words \citep{duan2024membership}, and 256 tokens \citep{shi2024detecting,Wang2024ConReCallDP}. Document-level MIA did not achieve significantly better results than random guessing \citep{meeus2024did,meeus_inherent_2024}, highlighting the challenges in scaling MIA techniques to larger text units. This gap between the granularity of current MIA methods and the document and collection of documents focus of legal disputes underscores the need for more comprehensive research into MIA effectiveness across different scales of textual data in LLMs.

%% file: sections/3-evaluation.tex
\begin{figure*}[t]
    \centering
    \includegraphics[width=0.9\linewidth]{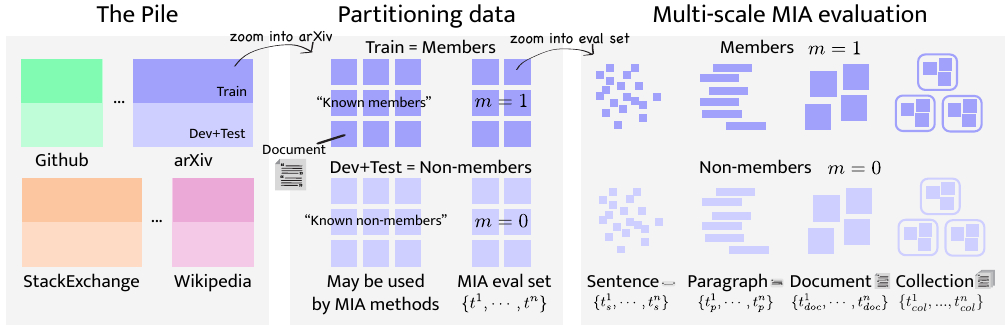}
    \caption{\textbf{Preparing MIA Evaluation Datasets.} Each source in The Pile is divided into Train, Dev, and Test splits, where the Train set is used for LLM training. For MIA, we designate the Train set as members and the Dev+Test sets as non-members. The MIA evaluation set ${t^1,\cdots,t^n}$ consists of datapoints for the binary detection task of predicting membership $m$. The benchmark makes some ``known members'' and ``known non-members'' available to the attackers; MIA methods may choose to use them or not. To support Collection-level MIA, we group documents in $n$ different ways to create member and non-member datasets.}
    \label{fig:data}
\end{figure*}

We introduce novel evaluation benchmarks for membership inference attacks (MIA) across various scales and multiple LLM training scenarios.

\paragraph{MIA Evaluation Overview.}
At a high level, following previous MIA work  \citep{shi2024detecting, meeus2024did, zhang_min-k_2024,wang_con-recall_2024, xie2024recall}, we frame the problem as a binary detection task: determining whether a given token sequence $t$ was used during the training of a target large language model $M$. Specifically, MIA methods are expected to produce scores $(s^1, \cdots, s^n)$ for each token sequence in a set of instances $(t^1, \cdots, t^n)$. With knowledge of the actual membership status $(m^1, \cdots, m^n)$, indicating whether each sequence was part of the training set, we can evaluate detection performance using the area under the receiver-operating-characteristic curve (AUROC) across different thresholds.

However, creating a binary detection task for long token sequences is non-trivial. Depending on the LLM type, sequences of 10K tokens or more often exceed the context window of current models. This requires assessing membership across a set of sequences, demanding datasets with known members and non-members at the sequence set level.

In this work, we propose four MIA benchmarks on the PILE dataset \citep{pile}, each targeting a different sequence length scale. We also extend previous MIA benchmarks, traditionally focused on detecting pre-training data, to three popular LLM training paradigms, including fine-tuning. In total, we introduce $4 \times 3 = 12$ MIA benchmarks, covering realistic application scenarios.

\begin{table}[t]
    \centering
    \small
    \setlength{\tabcolsep}{.35em}
    \begin{tabular}{@{}c@{\hskip .2em}lll@{}}
    \toprule
    \multicolumn{2}{l}{$\,\,$\textbf{Scale}} & \textbf{Definition} & \textbf{\#Tokens} \\
    \midrule
    \adjustbox{valign=m}{\includegraphics[width=2em]{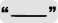}} & Sentence & Natural definition & 43 on avg. \\
    \rule{0pt}{1.7em} 
    \adjustbox{valign=m}{\includegraphics[width=2em]{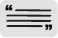}} & Paragraph & LLM context size & 512, 1024, 2048, $\ldots$ \\
    \rule{0pt}{1.7em} 
    \adjustbox{valign=m}{\includegraphics[width=2em]{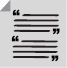}} & Document & Natural definition & 14K on avg. \\
    \rule{0pt}{1.7em} 
    \adjustbox{valign=m}{\includegraphics[width=2em]{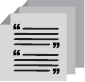}} & Collection & Multiple docs & 14K $\times$ \#docs \\
    \bottomrule
    \end{tabular}
    \vspace{-.5em}
    \caption{\textbf{MIA scales.} We define MIA at four scales.}
    \label{tab:mia-scales}
\end{table}

\paragraph{Data Scales.} We define four scales of the MIA tasks: i) sentence, ii) paragraph, iii) document, and iv) collection, as shown in \Cref{tab:mia-scales}.

\textbf{1) Sentence-level MIA.} We define a sentence as a natural sequence of words ending in a full stop. The average sentence in the Pile contains 43 tokens. This granularity is important because it is used to detect whether specific data points in benchmarks (e.g., questions in question-answering tasks) are contaminated, ensuring fair model evaluation. Due to its short nature, it can be extremely challenging to perform MIA successfully. For instance, \citet{duan2024membership} shows large overlaps between member and non-member sentences, which blurs the decision boundary. Additionally, sentence-level MIA can probe privacy leakage inferring membership of personally identifiable information \citep{kim2023_propile}. 

\textbf{2) Paragraph-level MIA.} We define a paragraph as a sequence of tokens that fits within the context window of a large language model (LLM). Thus, the length of a "paragraph" depends on the specific LLM in use. In our experiments, we use paragraphs of up to 512, 1024, and 2048 tokens, aligning with the context window sizes of the target models. In practice, paragraph-level copyrighted materials are particularly relevant to user-generated content on social media platforms and online forums, where short-form content is most common.

\textbf{3) Document-level MIA.} We define a document in the conventional sense, such as a single arXiv paper. In The Pile dataset, the average document contains 14,222 tokens. Documents typically consist of multiple paragraphs, often exceeding the context window of many LLMs, such as Pythia, which has a 2048-token limit. MIA approaches must handle these long documents by splitting the token sequence into smaller chunks, or ``paragraphs'' (as defined earlier), that fit within the model's context length and then aggregating the model’s responses across these chunks. Performing MIA at the document level is essential for addressing copyright and data ownership concerns, particularly for copyrighted materials like novels, news articles, research papers, and book chapters.

\textbf{4) Collection-level MIA.} We define a collection as a set of documents.
The attacker can choose the collection’s size. For instance, with 100 documents, a collection contains approximately 1.4 million tokens. Collection-level MIA is crucial, as one may question whether a collection of articles, such as those provided by an internet service provider, has been crawled and used in LLM training. Aggregating individual signals can amplify the detection of the collection’s usage. This is also relevant when examining contaminated benchmarks, as entire datasets may be unintentionally included in the training, leading to an overestimation of model performance.

MIA at scales beyond the document level is legally and practically significant, as copyright disputes often focus on individual articles. Moreover, as we will show in \S\ref{sec:experiments}, MIA only achieves strong performance at these larger scales.

\paragraph{LLM Training Scenarios.} Previous MIA approaches for LLMs have primarily focused on detecting pre-training data. While pre-training poses significant concerns regarding data scraping and exploitation by large tech companies with vast resources, other forms of training, such as fine-tuning, are much more common and affordable for entities worldwide. Although fine-tuning operates on a smaller scale, it may still involve copyrighted materials, and models fine-tuned in this way could be deployed in widely used applications. In such cases, copyright infringement could have serious consequences. Therefore, we consider MIA under three different LLM training scenarios.

\textbf{1) Pretraining MIA} requires the attacker to identify whether a given data instance was part of the pretraining corpus.

\textbf{2) Continual-learning MIA} addresses the challenge of keeping models up-to-date, as frequently retraining them from scratch is often infeasible due to high costs. Instead, continual learning offers a solution by incrementally training the model's checkpoint on new data. This approach is crucial for ensuring that LLMs remain current with evolving information \citep{wu2024continual}.

\textbf{3) Fine-tuning MIA} seeks to determine whether specific samples were used during fine-tuning for particular end tasks or domains, such as training data for conversational bots, question-answering systems, and similar applications. Fine-tuning methods are becoming increasingly popular due to their lower costs and flexibility, making them an ideal solution for tasks with large amounts of relevant data \citep{llama_community_stories}.

Beyond addressing various practical scenarios, our exploration of non-pretraining MIA highlights the increased effectiveness of existing MIA methods. Since fine-tuning is conducted on smaller datasets, each data instance has a more significant influence on model behaviour, which enhances the likelihood of successful MIA  (\S\ref{sec:experiments}).

\paragraph{Preparing Data for Multiscale MIA Evaluation.} 
For MIA evaluation, it is critical to have access to true members $m=1$ and true non-members $m=0$ for a precise evaluation of the membership detection performance. The Pile is one of the few datasets used for training LLMs that provides a clear separation of data used for training (the train split) and data \textit{not} used for training the LLMs (the dev and test splits). We thus re-purpose the Pile dataset for MIA evaluation.

\Cref{fig:data} illustrates the procedure of turning the Pile dataset into a MIA evaluation benchmark. For each source in the Pile, we use a subset of the train (members) documents and dev+test (non-members) documents for the MIA evaluation set $\{t^1,\cdots,t^n\}$. To enable multi-scale MIA evaluation, we adjust the granularity of each document instance $t^i$ by either breaking it down into sentences or paragraphs or by aggregating the documents into sets of documents (``collections'' in our definition). 

We encounter a problem at greater scales of MIA evaluation, especially for the collection-level MIA. Many sources of the Pile contain only a small number of documents as non-members. For instance, arXiv includes only 4.8K non-member documents. When the number of documents in each dataset is above 100, there are at most 48 available non-members to evaluate the MIA performance. To address this problem, we use bootstrapping \citep{efron1992bootstrap}: instead of considering only non-overlapping collections, allowing overlapping collections leads to $ 5 \times 10^{209} $ possible combinations of 100 documents. This also mirrors real-world scenarios, where overlapping documents are common across collections of documents, such as Wikipedia entries in question-answering datasets \citep{qa_dataset_explosion}. 
In our experiments, we generate 1K collections for evaluation for members and non-members and conduct experiments varying the number of documents in each collection between 10 and 500. 

The ``known members'' and ``known non-members'' splits are the ones that are not used for MIA evaluation. They may be used for improving or adapting MIA to the collection of interest. Our method in \S\ref{sec:method}, for example, uses the known non-members subset to calibrate the detection.

%% file: sections/4-methods.tex
We extend the Dataset Inference (DI) methods introduced by \citet{maini_llm_2024} to multiple MIA scales and adapt them for the binary detection task. While the original DI focused solely on determining whether a single dataset is a member or not, we adapt this methodology for finer granularity, such as document-level MIA. Additionally, we explain how to derive the detection scores necessary for evaluating AUROC performance.

We begin by explaining the Dataset Inference (DI) process.
Given a token sequence $t$, we divide it into smaller paragraphs, $t_{p,1}, \dots, t_{p,K}$, each small enough to fit within the context window of the large language model (LLM) $M$ under investigation. Following the DI approach, we compute various membership inference attack (MIA) features for each token sequence, using existing sentence-level MIA methods. These include perplexity \citep{yeom2018privacy}, lowercase perplexity, zlib compression scores \citep{carlini2021extracting}, and Min-K statistics (with thresholds at 5\%, 10\%, 20\%, ..., 60\%; \citealt{zhang_min-k_2024}). These MIA features, denoted by $A_1, \dots, A_L$, ($L = 10$) are functions that take the likelihood output of $M$ on a sequence as input and return a scalar representing the membership information for that sequence. Thus, for a given token sequence $t$, we produce a $K \times L$ array of MIA features, $A_l(M(t_{p,k}))$.

We employ a two-stage aggregation process to reduce the $K \times L$ array into a single statistic representing membership likelihood. In the first stage, we aggregate the $L$ MIA features into a single score using a learned linear map $f:\mathbb{R}^L \rightarrow [0,1]$. This linear map is trained on a dataset consisting of 1,000 known members and 1,000 known non-members (see \S\ref{sec:evaluation}), with the objective of predicting the membership status $m$ of the token sequence $t$.

During the second stage, we obtain a set of $K$ MIA scores: $\{f(t_{p,1}),\cdots,f(t_{p,K})\}$. These scores are treated as an unordered set, as the criteria for determining membership becomes largely independent of the ordering of documents within a dataset or paragraphs within a document. Following the DI approach, we perform statistical testing by comparing the set of $K$ MIA scores against a baseline set of $K$ scores from known non-members. Specifically, we use the Student's t-test for collection-level MIA since we aggregate more than 30 samples and the Mann–Whitney U-test for document-level MIA since we aggregate less than 30 samples from a non-normal distribution. 

\vspace{-1em}
{\small
\begin{align*}
    \text{t-score}(t) = \frac{\mu - \mu_\text{n-m}}{\sqrt{{s^2} + {s_\text{n-m}^2}}} \quad \\
    \text{U-score}(t) = - \sum_{k=1}^{K} \text{rank}(f(t_{p,k})) \nonumber
\end{align*}
}
\vspace{-1em}

Here, $\mu$ and $s$ represent the mean and sample standard deviation of the $K$ MIA scores, with the subscript n-m indicating the scores from the non-members.
The term $\text{rank} (f(t_{p,k}))$ refers to the rank (i.e., position in the list) of the MIA feature for the query sequence $t$ in the combined set of $2K$ MIA features, which includes both the query sequence and the known non-members.

We conduct all our experiments on Pythia 2.8B and 6.9B \citep{biderman2023pythia} with data from The Pile dataset \citep{pile}. We also provide the results of our main experiments on GPT Neo 2.7B in \Cref{appendix:mia_benchmark}.
See \Cref{app:setup} for details on the experimental setup.

%% file: sections/5-experiments.tex
With our experiments, we aim to answer the following questions:
i) How effective is the aggregation of MIA scores for larger textual units, such as documents and collection of documents? (\S\ref{sec:aggregation}),
ii) What are the requirements for a successful aggregation? (\S\ref{sec:aggregation_requirements}),
iii) What is the nature of the compounding effect observed in MIA score aggregation? (\S\ref{sec:compounding}),
iv) How much does the MIA aggregation benefit from fine-tuning scenarios (\S\ref{sec:fine_tuning_scenarios})? 

\begin{figure}[t]
\centering
\includegraphics[width=.8\linewidth]{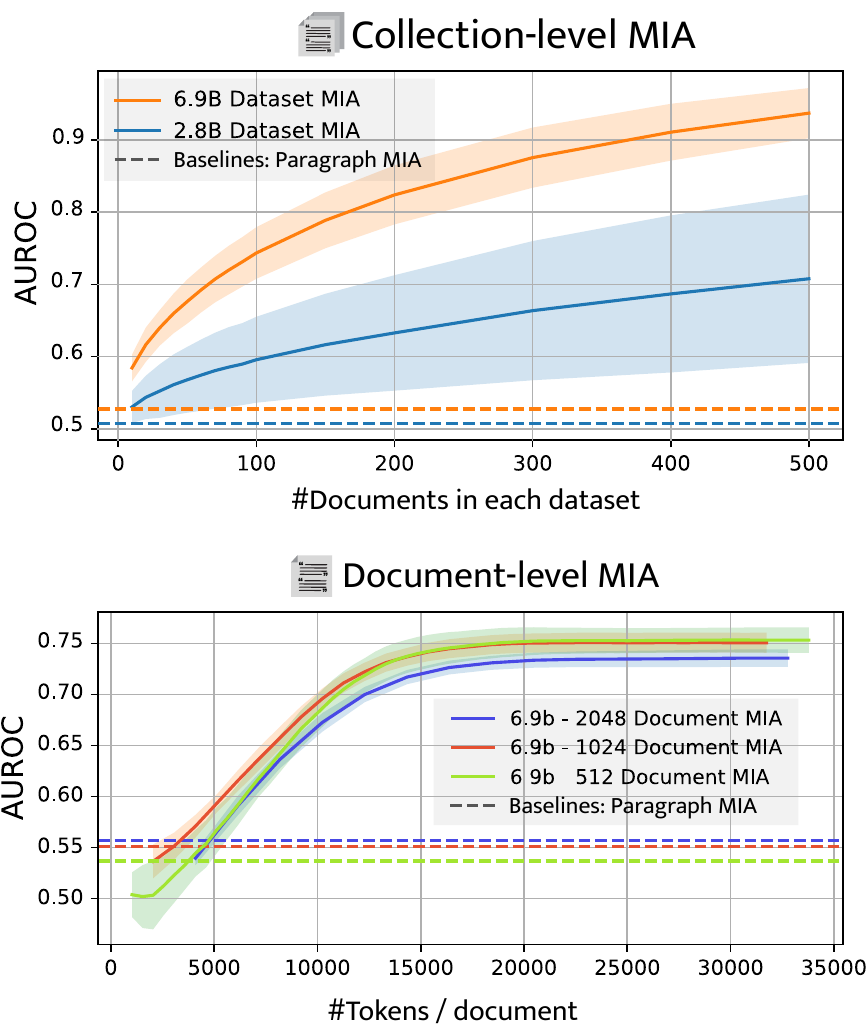}
\caption{\textbf{Effect of aggregation.} We show MIA performances on arXiv at different levels of aggregation. Aggregation becomes more effective as we increase the number of aggregated instances.}
\label{fig:aggr_working}
\end{figure}

\begin{figure*}[t]
\centering
\includegraphics[width=\textwidth]{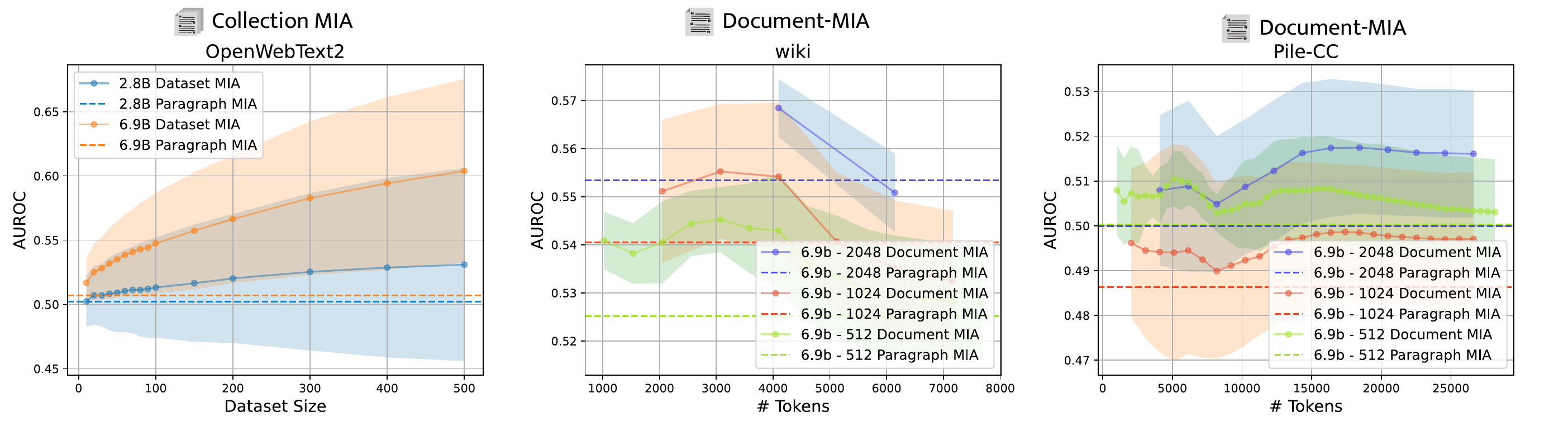}
\vspace{-.5em}
\caption{\textbf{Failute modes for aggregation.} Aggregation may not work when the base performance is too low (left and right plots), or when the amount of information to aggregate is too short (centre plot).}
\label{fig:aggr_not_working}
\end{figure*}

\subsection{Aggregating Text Subunits is Effective}
\label{sec:aggregation}
\input{sections/4-experiments/succesful_aggregation.tex}

\subsection{Requirements for Successful Aggregations}
\label{sec:aggregation_requirements}
\input{sections/4-experiments/aggregation_requirements.tex}

\subsection{Compounding Effect in MIA Score Aggregation}
\label{sec:compounding}
\input{sections/4-experiments/compounding.tex}

\subsection{MIA Aggregation Benefits from Fine-tuning Scenarios}
\label{sec:fine_tuning_scenarios}
\input{sections/4-experiments/finetuning.tex}

%% file: sections/4-experiments/succesful_aggregation.tex
\input{tables/pretrain}

\Cref{table:pretrain} and \Cref{fig:aggr_working} demonstrate the effectiveness of aggregating multiple text units (e.g., sentences or paragraphs) to perform successful Membership Inference Attacks (MIA) at larger textual scales (e.g., documents or collections). The figure illustrates a clear trend: MIA performance improves as we increase the number of aggregated text units. 

The upper plot focuses on collection MIA for arXiv. It reveals a stark contrast between small and large collections. While collections with only dozens of documents yield AUROC scores barely above random chance, those with 500 or more documents achieve significantly higher AUROC. Notably, using the 6.9B model and despite employing only a few MIA methods in our ensemble, we attain a remarkable collection MIA AUROC of 0.9 for arXiv.

The bottom plot shows the performance of document MIA on arXiv, where we observe our highest results. This exceptional performance can be attributed to two factors: i) the considerable length of the documents (averaging around 15K tokens), allowing for the aggregation of numerous MIA scores, and ii) a paragraph MIA AUROC exceeding 0.53. The combination of these factors yields an impressive document MIA AUROC of 0.75. To the best of our knowledge, this marks the first successful application of MIA to entire documents.

\Cref{table:mia_benchmark} in \Cref{appendix:mia_benchmark} includes all the results of the membership inference attacks (MIA) across all data levels for GPT Neo 2.7B, Pythia 2.8B, and Pythia 6.9B, showing that the trends holds for the three models.

\paragraph{Known Partition Size and MIA Performance}
We observe no correlation between the size of the known partition used in the statistical test and MIA performance. This finding is significant as it eliminates the need for large held-out collections, enhancing the method's practicality. Data providers often filter out portions of their collections during their cleaning process; we believe some of this filtered data could be used as the known non-members partition. This approach makes our methodology applicable to existing datasets without additional data collection. We provide a plot that compares the MIA performance across different known partition sizes in \Cref{fig:known_size} in \Cref{app:figures}.

%% file: tables/pretrain.tex
\begin{table}[t]
    \small
    \setlength{\tabcolsep}{.3em}
\centering
\begin{tabular}{lc@{}c@{}c@{}}
\toprule
\textbf{Data Scale} & \textbf{ArXiv}       & \textbf{HackerNews}  & \textbf{Wiki}   \\ \midrule
Sentence   &      0.501{\tiny ±0.003}      & 0.500{\tiny ±0.003}   & 0.507{\tiny ±0.004} \\
Paragraph  & 0.528{\tiny ±0.004} & 0.511{\tiny ±0.015} & 0.523{\tiny ±0.013} \\
Document   & 0.697{\tiny ±0.060} & 0.513{\tiny ±0.040} & 0.560{\tiny ±0.011} \\
Collection (500)    & 0.943{\tiny ±0.025} & 0.709{\tiny ±0.340} & 0.844{\tiny ±0.132} \\ \bottomrule
\end{tabular}
\caption{\textbf{Multi-scale MIA results.} We show the AUROC scores for Pythia 6.9B at four scales. For collection-level MIA, we use sets of 500 documents.}
\label{table:pretrain}
\end{table}

%% file: sections/4-experiments/aggregation_requirements.tex
\Cref{fig:aggr_not_working} illustrates common scenarios where the aggregation method fails to achieve MIA performance. The left plot demonstrates that collection-MIA can be ineffective when the base AUROC is close to random chance. Even with collection of 500 documents, the collection-MIA for the 2.8B model barely achieves an AUROC of 0.55. The central plot presents a case where the base AUROC is sufficiently high, but the documents are too short (consisting of only 3 paragraphs of 2048 tokens each), providing insufficient paragraphs for effective aggregation. The right plot shows the opposite situation: while the documents are long enough (comparable to the successful arXiv case), the base AUROC is only 0.5, rendering the aggregation ineffective. These examples highlight the importance of both base performance and sufficient text units for successful MIA score aggregation. Lastly, aggregating sentences to conduct paragraph MIA becomes extremely challenging because sentence-level MIA remains unachievable and the works of \citet{mainidataset, duan2024membership} suggest it might even be impossible. We provide the plots for all collection, document, and paragraph aggregations on \Cref{app:figures}.

%% file: sections/4-experiments/compounding.tex
\begin{figure}[t]
    \centering
    \includegraphics[width=\linewidth]{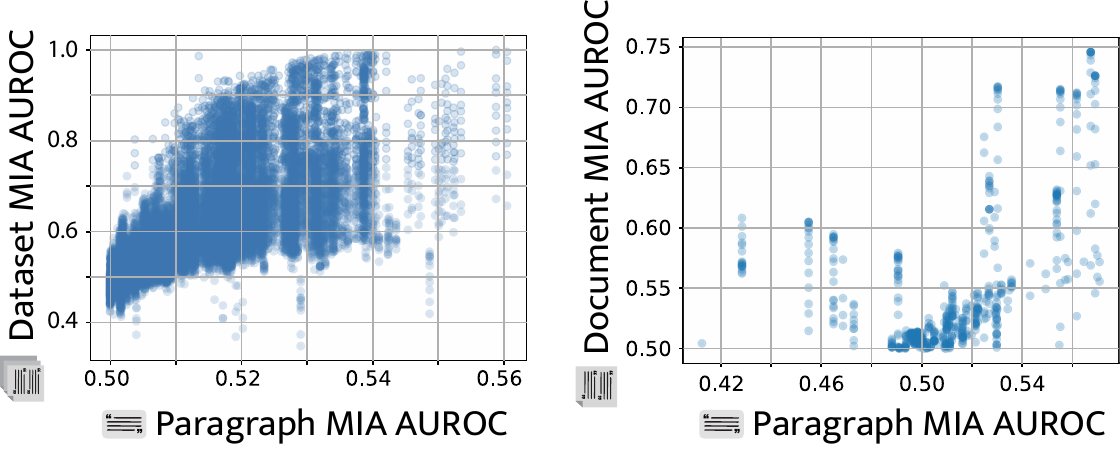}
    \caption{\textbf{Impact of paragraph-MIA performance} after aggregation to dataset- and document-level MIA.}
    \label{fig:compounding}
\end{figure}

Our experiments reveal a powerful compounding effect in the aggregation of MIA scores from smaller to larger textual units. This relationship follows an approximately square root function, where small improvements in paragraph-level MIA performance lead to substantial gains at the collection or document level. \Cref{fig:compounding} illustrates this relationship across all our experiments, encompassing various sources and hyperparameters.

The compounding effect is particularly striking when examining specific thresholds. For instance, a paragraph-level MIA with an AUROC of 0.51 can result in collection-level MIA AUROCs ranging from 0.5 to 0.65. However, a modest improvement in paragraph-level performance to an AUROC of 0.53 can dramatically boost collection-level AUROCs to between 0.6 and 0.9. This demonstrates the potential for significant gains in MIA effectiveness through relatively small improvements in paragraph level.

Lastly, our analysis also indicates that our aggregation method remains effective as long as the base AUROC exceeds 0.51 and establishes a threshold for when aggregation becomes a viable strategy.

%% file: sections/4-experiments/finetuning.tex
\subsubsection{Continual Learning Fine-tuning}

\input{tables/continual_learning}

\begin{figure}[t]
    \centering
    \includegraphics[width=.8\linewidth]{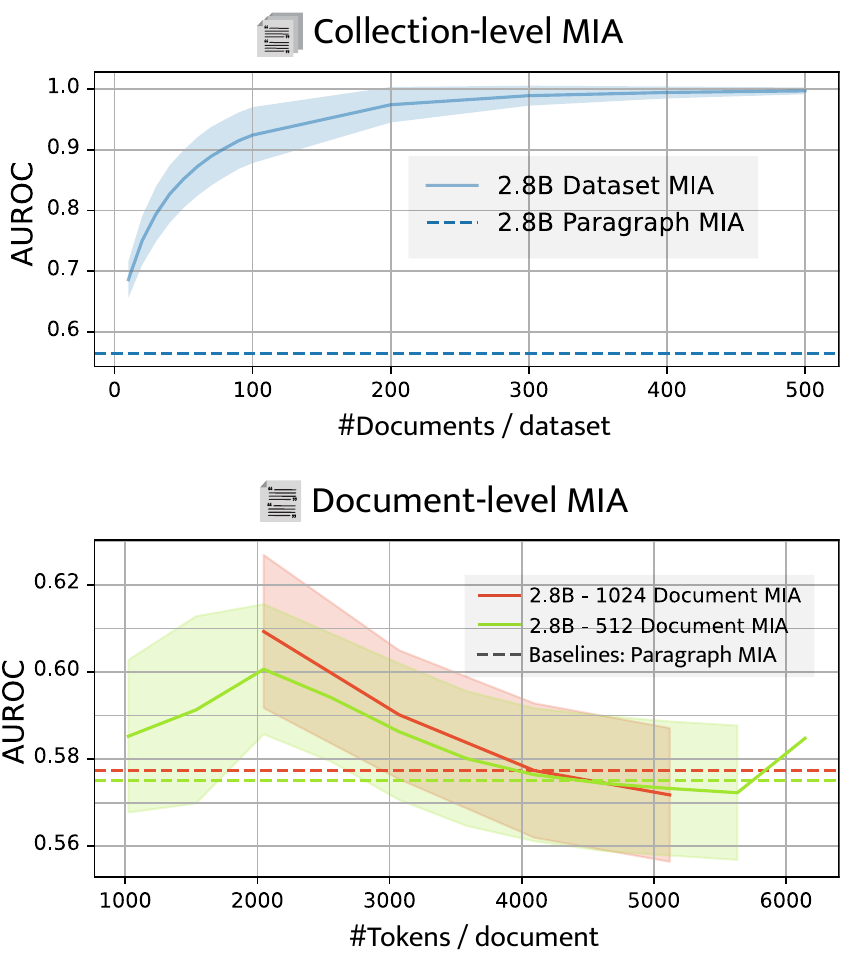}
    \vspace{-.5em}
    \caption{\textbf{Continual learning results} for Pythia 2.8B on Wikipedia. Collection-level MIA shows near-perfect AUROC with sufficient documents, while document-level MIA shows a significant improvement.}
    \label{fig:continual_indomain}
\end{figure}

In this section, we investigate the performance of MIA on LLMs that have undergone a continual learning process to adapt to specific domains. \Cref{fig:continual_indomain} and \Cref{table:ID} show the MIA performance on Pythia 2.8B after it was further trained on the validation sets of Wikipedia and GitHub (independently) from the Pile dataset.

Our results reveal a significant increase in MIA effectiveness in this continual learning scenario. For Wikipedia, the collection-level MIA performance achieves an AUROC of over 0.9, with collections containing only 100 documents. This stands in stark contrast to the pretraining scenario, where the 2.8B model only reached an AUROC of 0.65 for collections of 500 documents. This substantial improvement can be attributed to an increase in the paragraph MIA AUROC to 0.577, which amplifies the compounding effect (\S\ref{sec:compounding}). However, sentence-level MIA remains ineffective, and paragraph-level MIA remains below 0.6.

We observe a similar pattern for ArXiv and GitHub, suggesting a consistent trend across different domains. These findings lead us to conclude that while LLMs trained with continual learning remain robust against paragraph-level MIA, they become notably vulnerable to MIA when scores are aggregated across larger textual units.

\subsubsection{End-Task Fine-tuning}

We reuse the fine-tuned Phi-2 for reasoning from \citet{puerto2024finetuning} for our fine-tuning experiments. They fine-tuned the model on multiple question-answering datasets so that the model responded with a chain of thoughts. This task allows us to know if MIA could be used to evaluate the contamination of fine-tuned models at the sentence level, for example, if the model used the evaluation questions for training.

\Cref{tab:cot_mia} presents the performance of dataset and sentence-level MIA. For dataset-MIA, we use 100 datasets of 20 datasets each from 10k unique questions using a non-member known partition of 10 questions for the statistical comparison. For the evaluation of sentence-MIA, we use 5980 questions. In both cases, we run the experiments five times with different random seeds. Notably, sentence-MIA achieves an AUROC of 0.793 ± 0.024, while dataset-MIA is 0.99 for small datasets of just 20 data points. This suggests that MIA could serve as strong evidence in legal cases to prove the use of data for fine-tuning LLMs, in contrast to the claims made by \citet{zhang2024membershipinferenceattacksprove}.

We further explored scenarios where collections contain a mix of both member and non-member questions to evaluate the robustness of our collection-MIA method in the presence of noise. As shown in \Cref{fig:dataset_mi_mix_ft} in \Cref{app:figures}, with a 20\% contamination rate, the method continued to classify the collection as members, but at a contamination rate of 50\%, the method correctly assigned membership status in 50\% of cases. These results confirm the robustness of our approach to handling noisy collections.

\begin{table}[t]
\centering
\begin{tabular}{@{}lc@{}}
\toprule
\textbf{MIA} & \textbf{AUROC} \\ \midrule
Sentence     & 0.793 ± 0.024  \\
Collection (20) & 0.993 ± 0.012  \\ \bottomrule
\end{tabular}
\caption{\textbf{CoT fine-tuning.} Collection and sentence-level MIA results on CoT-fine-tuned Phi 2. Evaluation on 100 collections of 20 questions each from 10k unique questions.}
\label{tab:cot_mia}
\end{table}

%% file: tables/continual_learning.tex
\begin{table}[]
\centering
\setlength{\tabcolsep}{3pt}
\small
\begin{tabular}{lccc}
\toprule
\textbf{Data Scale}  & \textbf{ArXiv} & \textbf{Github}         & \textbf{Wikipedia}            \\ \midrule
Sentence   & 0.498{\tiny ± 0.004} & 0.496{\tiny ± 0.006} & 0.498{\tiny ± 0.006} \\
Paragraph  & 0.587{\tiny ± 0.009} & 0.559{\tiny ± 0.017} & 0.577{\tiny ± 0.012} \\
Document   & 0.582{\tiny ± 0.06}& 0.579{\tiny ± 0.014} & 0.590{\tiny ± 0.015} \\

Collection (500)   & 1.0{\tiny ± 0.0} & 0.885{\tiny ± 0.064} & 0.997{\tiny ± 0.007} \\ \bottomrule
\end{tabular}

\caption{\textbf{Multi-scale MIA for continual learning.} We report AUROC for a continually-trained Pythia 2.8B.}
\label{table:ID}
\end{table}

%% file: sections/6-conclusion.tex
In this paper, we have shown the importance of evaluating membership inference attacks (MIA) across different data scales, from sentences to collections of documents. Each text granularity represents a different and valuable use case that requires investigation. In contrast to prior works that suggest that MIA does not work for LLMs, we have empirically shown that MIA can work for document and collection levels with currently available attacks. We have further explored the performance of MIA across different training stages of an LLM and show that while small continual training remains robust towards sentence-level MIA, end-task fine-tuning is vulnerable, making MIA a suitable method to analyse test-set contamination.

In this work, we use simple MIA baselines to test the effectiveness of their aggregation and statistical testing. We believe the addition of more baselines would improve the overall results and leave that for future work.

%% file: sections/limitations.tex
The lack of training-validation-test set splits for training data of LLMs limits the range of models to evaluate. In this work, we only focus on Pythia as in \citep{duan2024membership, maini_llm_2024}. Furthermore, we only employed 2.8B and 6.9B, the smallest models where MIA has been seen to perform a bit better than random chance. Using larger models could boost the reported results. Similarly, we only use three baselines as membership inference attacks. The use of more baselines could further improve our results. 

%% file: sections/ethics.tex
This work adheres to the ACL Code of Ethics. In particular, The Pile and Pythia we used have been shown by prior works to be safe for research purposes. They are not known to contain personal information or harmful content. We also fulfill with their licenses MIT for The Pile and Apache 2.0 for Pythia. Similarly, the model used in \Cref{sec:fine_tuning_scenarios} also uses Apache 2.0 and is known to be safe for research purposes. Our method aims to provide data providers with tools to use their rights for the hypothetical cases where an LLM developer uses their data without consent.

Understanding when and how membership inference attacks (MIA) succeed on large language models (LLMs) can provide insights into developing strategies to train LLMs to be robust against these attacks. This poses the risk of making MIA outdated in the future, and therefore, data providers could be back in their current position without tools to use their rights.

%% file: tables/finetuning_hps.tex
\begin{table}[ht]
\centering
\begin{tabular}{@{}ll@{}}
\toprule
\textbf{Hyperparameter} & \textbf{Value}     \\ \midrule
Batch size              & 8                  \\
Epochs                  & 4                  \\
Block size              & 1024               \\
Optimizer               & Paged 32-bit AdamW \\
Learning rate           & $2\times10^{-4}$     \\
Gradient clipping norm  & 0.3                \\
Weight decay            & 0.001              \\
Quantization            & 4 bits             \\
LoRa $\alpha$           & 16                 \\
LoRa rank $r$           & 64                 \\
LoRa dropout            & 0.1                \\ \bottomrule
\end{tabular}
\caption{Hyperparameters used to fine-tune the continual learning models.}
\label{table:ft-hps}
\end{table}

%% file: tables/u-test.tex
\begin{table}[t]
\begin{tabular}{@{}lcc@{}}
\toprule
\textbf{Dataset}                    & \textbf{Method} & \textbf{AUROC}         \\ \midrule
\multirow{2}{*}{ArXiv}              & t-test          & 0.943 ± 0.025          \\
                                    & u-test          & \textbf{0.951 ± 0.026} \\\midrule
\multirow{2}{*}{FreeLaw}            & t-test          & 0.551 ± 0.126          \\
                                    & u-test          & \textbf{0.594 ± 0.206} \\\midrule
\multirow{2}{*}{Github}             & t-test          & 0.497 ± 0.088          \\
                                    & u-test          & \textbf{0.57 ± 0.162}  \\\midrule
\multirow{2}{*}{HackerNews}         & t-test          & \textbf{0.709 ± 0.34}  \\
                                    & u-test          & 0.702 ± 0.253          \\\midrule
\multirow{2}{*}{OpenWebText2}       & t-test          & 0.615 ± 0.089          \\
                                    & u-test          & \textbf{0.651 ± 0.113} \\\midrule
\multirow{2}{*}{Pile-CC}            & t-test          & \textbf{0.55 ± 0.132}  \\
                                    & u-test          & 0.549 ± 0.155          \\\midrule
\multirow{2}{*}{PubMed\_Abstracts}  & t-test          & 0.637 ± 0.056          \\
                                    & u-test          & \textbf{0.659 ± 0.094} \\\midrule
\multirow{2}{*}{StackExchange}      & t-test          & 0.693 ± 0.175          \\
                                    & u-test          & \textbf{0.73 ± 0.17}   \\\midrule
\multirow{2}{*}{USPTO\_Backgrounds} & t-test          & 0.609 ± 0.151          \\
                                    & u-test          & \textbf{0.652 ± 0.189} \\ \midrule
\multirow{2}{*}{Wikipedia}          & t-test          & 0.85 ± 0.144           \\
                                    & u-test          & 0.855 ± 0.171          \\ \bottomrule
\end{tabular}
\caption{Comparison between t-test and u-test on Dataset MIA.}
\label{tab:u-test}
\end{table}

%% file: tables/mia_benchmark.tex
\begin{table*}[]
\centering
\begin{tabular}{@{}llcccc@{}}
\toprule
\textbf{Dataset}                    & \textbf{Model} & \textbf{Collection} & \textbf{Document} & \textbf{Paragraph} & \textbf{Sentence} \\ \midrule
\multirow{3}{*}{ArXiv}              & GPT-Neo 2.7B  & \textbf{0.608 ± 0.
163}     & 0.522 ± 0.011       & 0.504 ± 0.006         & 0.497 ± 0.002      \\
                                    & Pythia 2.8B  & \textbf{0.718 ± 0.122}     & 0.523 ± 0.01       & 0.509 ± 0.006       & 0.499 ± 0.003      \\
                                    & Pythia 6.9B  & \textbf{0.943 ± 0.025}     & \textbf{0.697 ± 0.06  }     & 0.557 ± 0.008       & 0.501 ± 0.003      \\\midrule
\multirow{3}{*}{FreeLaw}           & GPT-Neo 2.7B  & 0.483 ± 0.148     & 0.522 ± 0.011       & 0.498 ± 0.008         & 0.506 ± 0.002      \\
                                    & Pythia 2.8B  & 0.511 ± 0.106     & 0.51 ± 0.017       & 0.509 ± 0.014       & 0.504 ± 0.002      \\
                                    & Pythia 6.9B  & 0.551 ± 0.126     & 0.538 ± 0.024      & 0.526 ± 0.011       & 0.505 ± 0.001      \\ \midrule
\multirow{3}{*}{Github}             & GPT-Neo 2.7B  & 0.496 ± 0.058      & 0.500 ± 0.008        & 0.498 ± 0.003         & 0.497 ± 0.004     \\
                                    & Pythia 2.8B  & 0.479 ± 0.069     & 0.498 ± 0.01       & 0.494 ± 0.009       & 0.496 ± 0.006      \\
                                    & Pythia 6.9B  & 0.497 ± 0.088     & 0.491 ± 0.005      & 0.494 ± 0.006       & 0.865 ± 0.004      \\
                                    \midrule
\multirow{3}{*}{HackerNews}         & GPT-Neo 2.7B  & \textbf{0.774 ± 0.077}      & 0.511 ± 0.027       & 0.509 ± 0.007         & 0.501 ± 0.002     \\
                                    & Pythia 2.8B  & \textbf{0.686 ± 0.267}     & 0.488 ± 0.02       & 0.511 ± 0.025       & 0.498 ± 0.002      \\
                                    & Pythia 6.9B  & \textbf{0.709 ± 0.34}      & 0.513 ± 0.04       & 0.514 ± 0.018       & 0.5 ± 0.003        \\
                                    \midrule
\multirow{3}{*}{OpenWebText2}       & GPT-Neo 2.7B  & 0.518 ± 0.078      & 0.498 ± 0.015       & 0.5010 ± 0.003         & 0.497 ± 0.006      \\
                                    & Pythia 2.8B  & 0.544 ± 0.097     & 0.505 ± 0.009      & 0.502 ± 0.008       & 0.497 ± 0.003      \\
                                    & Pythia 6.9B  & \textbf{0.615 ± 0.089 }    & 0.519 ± 0.008      & 0.513 ± 0.006       & 0.499 ± 0.005      \\\midrule
\multirow{3}{*}{Pile-CC}            & GPT-Neo 2.7B  & 0.516±0.089   & 0.497 ± 0.007        & 0.501 ±0.006        & 0.499 ± 0.007       \\
                                    & Pythia 2.8B  & 0.52 ± 0.095      & 0.489 ± 0.011      & 0.481 ± 0.005       & 0.495 ± 0.004      \\
                                    & Pythia 6.9B  & 0.55 ± 0.132      & 0.513 ± 0.015      & 0.5 ± 0.013         & 0.931 ± 0.005      \\\midrule
\multirow{3}{*}{USPTO} & GPT-Neo 2.7B  & 0.501	±0.066      & 0.482±0.007        & 0.499±0.004         & 0.496±0.002      \\
                                    & Pythia 2.8B  & 0.512 ± 0.112     & 0.495 ± 0.032      & 0.49 ± 0.014        & 0.5 ± 0.001        \\
                                    & Pythia 6.9B  & \textbf{0.609 ± 0.151}     & 0.534 ± 0.007      & 0.52 ± 0.005        & 0.497 ± 0.002      \\ \midrule
\multirow{3}{*}{Wikipedia}          & GPT-Neo 2.7B & \textbf{0.648 ± 0.103}     & 0.517 ± 0.017       & 0.508 ± 0.006         & 0.497 ± 0.005     \\
                                    & Pythia 2.8B  & \textbf{0.665 ± 0.169}     & 0.531 ± 0.019      & 0.534 ± 0.015       & 0.503 ± 0.006  \\
                                    & Pythia 6.9B  & \textbf{0.85 ± 0.144 }     & 0.56 ± 0.011       & 0.553 ± 0.01        & 0.507 ± 0.004      \\ \bottomrule

\end{tabular}
\caption{AUROC performance of Membership Inference Attacks across data scales. Results $> 0.6$ bolded.}
\label{table:mia_benchmark}
\end{table*}

%% file: figures/tab_nb_tokens_datasets.tex
\begin{table*}[ht]
\centering
    \scalebox{0.95}{
\begin{tabular}{l@{}r@{}l@{}r@{}l@{}r@{}l@{}r@{}l@{}r@{}lr@{}l}
\toprule
 & \multicolumn{2}{c}{} & \multicolumn{6}{c}{Paragraph} & \multicolumn{2}{c}{} & \multicolumn{2}{c}{} \\
   \cmidrule(lr){4-9} 
 ThePile Subset & \multicolumn{2}{c}{Sentence} & \multicolumn{2}{c}{\small 512 Tokens} & \multicolumn{2}{c}{\small 1024 Tokens} & \multicolumn{2}{c}{\small 2048 Tokens} & \multicolumn{2}{c}{Document} & \multicolumn{2}{c}{Collection \small{(500 Docs)}} \\
\midrule
ArXiv & 43.3 & {\tiny $\pm$ 43.0} & 497.7 & {\tiny $\pm$ 73.3} & 984.4 & {\tiny $\pm$ 177.0} & 1948.5 & {\tiny $\pm$ 403.2} & 14221.9 & {\tiny $\pm$ 13893.0} & 7102765.9 & {\tiny $\pm$ 275483.6} \\
DM Maths & 23.0 & {\tiny $\pm$ 18.8} & 512.0 & {\tiny $\pm$ 0.0} & 1024.0 & {\tiny $\pm$ 0.0} & 2048.0 & {\tiny $\pm$ 0.0} & 3693.9 & {\tiny $\pm$ 652.9} & 1846939.7 & {\tiny $\pm$ 12665.8} \\
GitHub & 127.1 & {\tiny $\pm$ 275.9} & 372.1 & {\tiny $\pm$ 178.0} & 590.0 & {\tiny $\pm$ 385.8} & 857.0 & {\tiny $\pm$ 747.0} & 1765.9 & {\tiny $\pm$ 3652.0} & 880755.8 & {\tiny $\pm$ 81174.1} \\
HackerNews & 30.8 & {\tiny $\pm$ 27.4} & 333.6 & {\tiny $\pm$ 182.6} & 527.8 & {\tiny $\pm$ 399.0} & 793.9 & {\tiny $\pm$ 778.6} & 1574.6 & {\tiny $\pm$ 7769.4} & 784320.5 & {\tiny $\pm$ 142487.2} \\
OpenWebText2 & 31.9 & {\tiny $\pm$ 37.4} & 392.6 & {\tiny $\pm$ 166.0} & 595.0 & {\tiny $\pm$ 356.1} & 754.7 & {\tiny $\pm$ 608.3} & 922.7 & {\tiny $\pm$ 1403.4} & 461230.0 & {\tiny $\pm$ 29294.9} \\
Pile-CC & 28.3 & {\tiny $\pm$ 24.6} & 363.5 & {\tiny $\pm$ 170.3} & 536.1 & {\tiny $\pm$ 359.3} & 688.9 & {\tiny $\pm$ 622.4} & 1011.5 & {\tiny $\pm$ 2770.4} & 503924.8 & {\tiny $\pm$ 57360.4} \\
PubMed Central & 38.9 & {\tiny $\pm$ 38.0} & 466.7 & {\tiny $\pm$ 135.1} & 913.0 & {\tiny $\pm$ 294.5} & 1763.0 & {\tiny $\pm$ 642.7} & 7124.0 & {\tiny $\pm$ 6825.5} & 3560020.9 & {\tiny $\pm$ 145890.2} \\
StackExchange & 53.0 & {\tiny $\pm$ 109.3} & 393.8 & {\tiny $\pm$ 132.0} & 519.8 & {\tiny $\pm$ 289.3} & 586.5 & {\tiny $\pm$ 444.6} & 627.2 & {\tiny $\pm$ 685.5} & 314027.1 & {\tiny $\pm$ 15372.1} \\
Wikipedia & 34.2 & {\tiny $\pm$ 44.3} & 323.3 & {\tiny $\pm$ 176.8} & 452.5 & {\tiny $\pm$ 348.4} & 564.5 & {\tiny $\pm$ 577.9} & 726.6 & {\tiny $\pm$ 1514.2} & 363979.1 & {\tiny $\pm$ 33328.6} \\
\bottomrule
\end{tabular}
}
\caption{\textbf{Number of tokens on the data scale}. Average and standard deviation of the number of tokens computed for several text granularities: sentence (sentences longer than 25 characters in the first 2048 tokens), paragraph (the first chunk of 512, 1024 and 2048 tokens), full document, collection (a set of 100 documents).}
\label{tab:token-metrics}
\end{table*}